\documentclass[letterpaper]{article} 
\usepackage{aaai24}  
\usepackage{times}  
\usepackage{helvet}  
\usepackage{courier}  
\usepackage[hyphens]{url}  
\usepackage{graphicx} 
\usepackage{natbib}  
\usepackage{caption} 
\usepackage{adjustbox}
\usepackage{amssymb}
\usepackage{algorithm}
\usepackage{fancyhdr}
\usepackage{algpseudocode}
\usepackage{multirow}
\usepackage{makecell}
\usepackage{amsmath}

\usepackage{newfloat}
\usepackage{listings}
\DeclareCaptionStyle{ruled}{labelfont=normalfont,labelsep=colon,strut=off} 
\floatstyle{ruled}
\newfloat{listing}{tb}{lst}{}
\floatname{listing}{Listing}
\title{Boosting Adversarial Transferability across Model Genus by Deformation-Constrained Warping}
\author{
        Qinliang Lin\textsuperscript{\rm 1}\equalcontrib, Cheng Luo\textsuperscript{\rm 1}\equalcontrib, Zenghao Niu\textsuperscript{\rm 1}, Xilin He\textsuperscript{\rm 1}, Weicheng Xie\textsuperscript{\rm 1, \rm 2, \rm 3}\thanks{Corresponding author.}\\
        Yuanbo Hou\textsuperscript{\rm 4}, Linlin Shen\textsuperscript{\rm 1, \rm2, \rm 3}, Siyang Song\textsuperscript{\rm 5}\\
}
\affiliations{
    \textsuperscript{\rm 1}Computer Vision Institute, School of Computer Science \& Software Engineering, Shenzhen University\\
    \textsuperscript{\rm 2}Shenzhen Institute of Artificial Intelligence and Robotics for Society\\
    \textsuperscript{\rm 3}Guangdong Key Laboratory of Intelligent Information Processing\\
    \textsuperscript{\rm 4}WAVES Research Group, Ghent University, Belgium\\
    \textsuperscript{\rm 5}University of Leicester, UK.\\
    2017192020@email.szu.edu.cn, wcxie@szu.edu.cn

}

\begin{document}

\maketitle

\begin{abstract}

Adversarial examples generated by a surrogate model typically exhibit limited transferability to unknown target systems.
To address this problem, many transferability enhancement approaches (\emph{e.g.,} input transformation and model augmentation) have been proposed. However, they show poor performances in attacking systems having different model genera from the surrogate model.
In this paper, we propose a novel and generic attacking strategy, called Deformation-Constrained Warping Attack (DeCoWA), that can be effectively applied to cross model genus attack.
Specifically, DeCoWA firstly augments input examples via an elastic deformation, namely Deformation-Constrained Warping (DeCoW), to obtain rich local details of the augmented input.
To avoid severe distortion of global semantics led by random deformation, DeCoW further constrains the strength and direction of the warping transformation by a novel adaptive control strategy.
Extensive experiments demonstrate that the transferable examples crafted by our DeCoWA on CNN surrogates can significantly hinder the performance of Transformers (and vice versa) on various tasks, including image classification, video action recognition, and audio recognition. Code is made
available at \url{https://github.com/LinQinLiang/DeCoWA}.
\end{abstract}

\section{Introduction}
In the past decade, various deep network architectures, including Convolutional Neural Networks (CNNs) \cite{DBLP:conf/cvpr/HeZRS16}, LSTMs \cite{hochreiter1997long}, Transformers~\cite{DBLP:conf/iclr/DosovitskiyB0WZ21}, etc., have demonstrated revolutionary performances in recognition or classification of real-world signals, including images~\cite{DBLP:conf/cvpr/HeZRS16}, videos~\cite{DBLP:conf/icml/BertasiusWT21}, audio~\cite{DBLP:journals/corr/abs-2210-11407}, etc. 
\begin{figure}[!tb]
     \centering
    \footnotesize
    \includegraphics[width=1.0\linewidth]{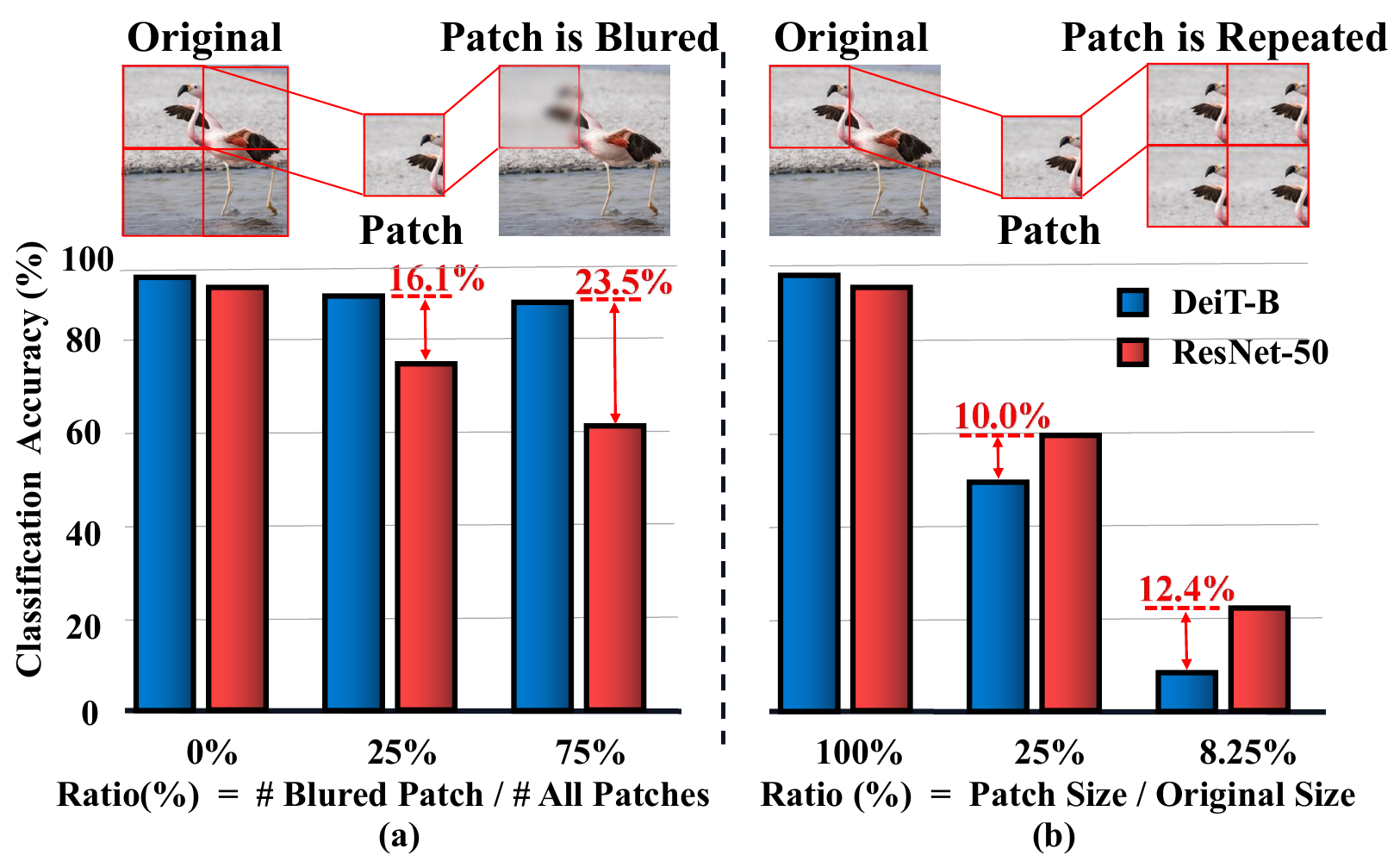}
  \caption{Illustration of distinctions between CNNs and ViTs. We test recognition accuracy by two different model genera using (a) local detail blurred images and (b) global structure damaged images, respectively.
  (a) As more local details are blurred, the performance of ResNet-50 significantly drops while DeiT-B is still robust. (b) When an image patch of a smaller size remains, ResNet-50 achieves higher classification accuracy than DeiT-B. \# denotes counting the number.}
     \label{fig:motivation}

 \end{figure}
Nevertheless, almost all these networks are vulnerable to adversarial examples crafted by adversaries, potentially leading to severe security threats. To this end, investigating generic features/vulnerabilities among different network architectures is crucial and urgent. 

Some approaches aim to generate adversarial examples with transferability to fool target systems in black-box scenarios, where network details of the target model are unknown.
Most of the relevant research~\cite{DBLP:conf/cvpr/DongLPS0HL18, DBLP:conf/iclr/LinS00H20, DBLP:conf/cvpr/DongPSZ19, DBLP:conf/cvpr/Wang021} lies on adversarial transferability across CNN architectures.
Meanwhile, a small part of recent studies~\cite{DBLP:conf/aaai/WeiCGWGJ22, DBLP:conf/iclr/NaseerR0KP22, wang2022generating, DBLP:journals/nca/HanLLJGGC22} has sparked a discussion about boosting the transferability across different Vision Transformers (ViTs).
Despite the above advances, there is still a lack of effective attacking methods that can achieve strong transferability in a wider and more practical setting (\emph{i.e.,} transferability across \textbf{model genus\footnote{One Model Genus is defined as a set of deep neural networks that have similar architectures. This concept was introduced in \cite{DBLP:conf/iccv/MahmoodMD21}, existing models were categorized into several model genera, such as ViT model genus, and CNN model genus.}}). 

To the best of our knowledge, this is one of the pioneering works aiming at improving 
\textbf{cross-model genera attacking transferability} (\emph{e.g.,} the surrogate model is CNN, whereas the target system is based on a Transformer model).
This task is more challenging due to the huge architecture gap between different model genera and the consequent distinctions of their extracted features \cite{DBLP:conf/nips/NaseerRKHKY21, DBLP:conf/nips/RaghuUKZD21, shao2021adversarial}.
Hereby, we present a few toy experiments to reveal the key distinction between discriminative features learned by CNN and ViT, respectively.

(i) As shown in Figure~\ref{fig:motivation}(a), we divide each image into several equal-sized patches and randomly blur some of them. It can be seen that with more patches blurred, pre-trained ResNet-50 \cite{DBLP:conf/cvpr/HeZRS16} achieves lower classification accuracy, whereas \textbf{DeiT-B} \cite{touvron2021training} is \textbf{more robust to local detail loss}. 
(ii) Figure~\ref{fig:motivation}(b) shows that when we pick one of the partitioned patches to fill the whole image so as to retain only local image features as input to classification models, the results demonstrate that \textbf{ResNet-50} can still \textbf{utilize local details as classification cues} and achieve relatively higher accuracy. 
These toy experiments suggest that both Transformer and CNN models make predictions mainly based on global features, {while} CNN models frequently {overfit} to local patterns. This also inspired us to pay attention to algorithms that are perceptive to local characteristics in images.

During transferable adversarial attacks, to the best of our knowledge, cross model genus attacks have not been 
{well explored}
in depth. Although model augmentation methods~\cite{DBLP:conf/iclr/LinS00H20} is a kind of effective solution that can simulate the decision space of similar architectures by input transformation. However, existing model augmentation methods fail to enhance transferability across model genera.
The reason is that they are specifically designed for the unique properties of a model genus and neglect the general and invariant features of different model genera (\textbf{Problem 1}).
For example, scale~\cite{DBLP:conf/iclr/LinS00H20} and translation invariance~\cite{DBLP:conf/cvpr/DongPSZ19} of convolutional operation is utilized to produce augmented samples, which can generalize the feature space of a CNN surrogate to another of the same model genus. 
However, these augmentation methods bring less benefit to the generalization to a different model genus (\emph{e.g.,} ViTs).
Furthermore, previous works such as affine/linear (\emph{e.g.,}, Admix~\cite{DBLP:conf/iccv/WangH0021}), intensity (\emph{e.g.,} SI~\cite{DBLP:conf/iclr/LinS00H20}) and spectrum (\emph{e.g.,} S$^2$I~\cite{DBLP:conf/eccv/LongZZGLZS22}) transformation techniques, have primarily focused on global contents, neglecting the importance of diverse local regions (\textbf{Problem 2}).
Hence, based on the aforementioned analysis, from the perspective of input transformation, we found that elastic deformation can adjust the local shape and contents while obtaining more augmented local patterns, thus extracting more local the general features.

Consequently, we propose a novel and generic input transformation approach called \textbf{Deformation-Constrained Warping (DeCoW)}, which applies a deformation to the local details of the target data (\emph{e.g.,} an image and log mel spectrogram of an audio segment).
In this way, the surrogate model has a tendency to rely on invariant features (\emph{i.e.,} global features) from augmented samples.
However, unconstrained elastic deformation may lead to excessive or unreasonable changes in image semantics.
To tackle this issue, we further contribute an adaptive control strategy in DeCoW which can optimize the random deformation variable to a constrained point, ensuring the consistency of the global semantics of augmented samples and inputs. The main contributions and novelties of this paper can be summarised as
\begin{itemize}
    \item We systematically investigated the task of cross model genus attack and revealed that the low transferability of adversarial samples generated by previous attack methods is due to insufficient manipulation on the local structure of the signal.

    \item We propose a generic Deformation-Constrained Warping~(DeCoW) to boost the adversarial transferability across model genera. DeCoW can increase the diversity of local details such as local shape and contours through elastic deformation and adaptively constrains the magnitude and direction of warping transformation.

    \item Deformation-Constrained Warping Attack~(DeCoWA) is proposed by integrating DeCoW into a gradient-based attack method. It is an approach that can be applied to various modalities of data such as image, video, and audio, and achieves superior transferability over the state-of-the-art attack methods by a significant margin.

\end{itemize}
\section{Related Work}
\subsection{Intriguing Properties of CNN and Transformer}
It is noted that CNNs excels at capturing high-frequency components of an image \cite{DBLP:conf/cvpr/WangWHX20} since their primary parts, convolutional layers, function as individual high-pass filters \cite{DBLP:conf/iclr/ParkK22}.
By contrast, Multi-Head Self-Attentions (MSAs), which bring key benefits to a transformer, serve as low-pass filters for spatial smoothing \cite{DBLP:conf/iclr/ParkK22}.
This difference between the main components of these two model genera unveils why CNNs are more sensitive to trivial details and noise as compared to ViTs. 
Furthermore, ViTs are confirmed to have a stronger bias to object shape than CNNs, whereas CNNs show more bias to local textures \cite{DBLP:conf/nips/NaseerRKHKY21, DBLP:conf/iclr/GeirhosRMBWB19}.
From a different perspective, Raghu \emph{et al.} \cite{raghu2021vision} leveraged Centered Kernel Alignment (CKA) to measure the layer representation similarity between network blocks. They found that ViTs have more consistent representations across all layers than CNNs.
These works expose different properties of the two model genera (\emph{i.e.,} CNN and ViT), which are beneficial to get out of the predicament of low adversarial transferability across different genera.

\subsection{{Deformation-based Data Augmentation}}

Data augmentation is one of the most valid approaches to boost model generalization. For instance, some model-free transformations like affine transformation (\emph{e.g.} scale, translation) and intensity transformation (\emph{e.g.} blurring and adding noise) are commonly used for training various models~\cite{DBLP:conf/cvpr/HeZRS16, DBLP:conf/cvpr/He0WXG20,DBLP:conf/nips/SohnBCZZRCKL20}. In addition to these methods, blending multiple images such as mixup~\cite{DBLP:conf/iclr/ZhangCDL18} have also been proposed to improve model generalization. Elastic deformation is another approach that is gradually being applied to model training because it can alter the shape or posture of objects~\cite{DBLP:journals/pr/XuYFP23}. 

\subsection{Transferable Adversarial Attack} \label{sec:TAA}

{\bf Transferable Attack on CNNs.} 
For transferable attacks across different CNN models, there is a large body of works spanning from gradient-based enhanced methods \cite{DBLP:conf/cvpr/DongLPS0HL18, DBLP:conf/iclr/LinS00H20, DBLP:conf/bmvc/WangL00021, DBLP:conf/cvpr/WeiCWJ22, DBLP:conf/aaai/WeiCWJCZJ20}, variance-tuning methods \cite{DBLP:conf/cvpr/Wang021, DBLP:conf/cvpr/XiongLZH022}, knowledge-based methods \cite{gao2021push, wu2020boosting, DBLP:conf/iccv/GaneshanSR19, huang2019enhancing, inkawhich2019feature, wang2021feature, zhou2018transferable, DBLP:conf/eccv/LongZZGLZS22, DBLP:conf/cvpr/LuoL0WXS22}, methods built on generative models \cite{DBLP:conf/nips/NakkaS21,DBLP:conf/eccv/YangDPSZ22}, to input augmentation methods \cite{DBLP:conf/iclr/LinS00H20, DBLP:conf/cvpr/XieZZBWRY19, DBLP:conf/cvpr/DongPSZ19, DBLP:conf/iccv/WangH0021}.
Among these, MIM \cite{DBLP:conf/cvpr/DongLPS0HL18}, a gradient-based enhanced method, uses a momentum term to keep the gradient directions, and SI-NI-FGSM \cite{DBLP:conf/iclr/LinS00H20}
further modifies it with a Nesterov momentum and variance tuning.
Recent research points to the great potential of augmentation techniques to boost adversarial transferability. 
For example, differentiable stochastic transformations, DI-FGSM \cite{DBLP:conf/cvpr/XieZZBWRY19}, is applied to aggregate diverse directions of gradients, Admix \cite{DBLP:conf/iccv/WangH0021} incorporates a small portion of images of other classes into the input example to increase gradient diversity, and S$^{2}$I-FGSM \cite{DBLP:conf/eccv/LongZZGLZS22} augments input from a perspective of the frequency domain. \cite{zhang2023improving} constructed a candidate augmentation path pool to augment images from multiple image augmentation paths. And \cite{liang2023styless} advocated using stylized networks to prevent adversarial examples from using non-robust style features.

\noindent {\bf Transferable Attack on Transformers.}
The popularity of ViT incurs a rapidly expanding body of studies on developing effective algorithms to improve the transferability of adversarial examples across various ViT models.
Naseer \emph{et al.} \cite{DBLP:conf/iclr/NaseerR0KP22} propose self-ensemble (SE) and token refinement (TR) strategies to alleviate the sub-optimal results.
However, Wei \emph{et al.}~\cite{DBLP:conf/aaai/WeiCGWGJ22} point out that there are not enough class tokens to construct an SE in the real world. 
To avoid it, the skills of Pay No Attention (PNA) and PatchOut are designed to manipulate the attention mechanism and image features in parallel.
Besides, Han \emph{et al.}~\cite{DBLP:journals/nca/HanLLJGGC22} also utilize partial encoder blocks to replace all encoder blocks, effectively reducing overfitting to a specific surrogate model.
Wang \emph{et al.}~\cite{wang2022generating} proposed an Architecture-Oriented Transferable Attacking (ATA) framework to take the architectural features of vision transformers into account. 
Zhang \emph{et al.}~\cite{zhang2023transferable} proposed the Token Gradient Regularization (TGR) to reduce the variance of the back-propagated gradient and utilizes the regularized gradient to generate adversarial samples.

\section{Methodology}

\subsection{Preliminary and A Unified Paradigm} \label{sec:pre_and_uni}
Formally, take the image classification model as an example, let $\mathcal{M}_{\phi}:x\rightarrow y $ 
represents a classifier with learned parameters $\phi$, $x \in \mathbb{R}^{H \times W \times C}$ and $y \in \mathcal{Y} = \{1, 2,.., \#class\}$ denotes clean input and ground truth, respectively, where $\#class$ represents the number of classes. Unlike the unconstrained spatial transformation perturbation \cite{xiao2018spatially}, our goal is to craft an adversarial example $x^{adv} = x + \delta$ with perturbations $\delta$, which can mislead the classifier to make a wrong decision, \emph{i.e.}, $\mathcal{M}_{\phi}(x^{adv}) \neq y$ (untargeted attack).
To limit attack strength, $x^{adv}$ is supposed to be constrained in a $\ell_{p}$-norm ball centered at $x$ with a radius $\epsilon$. 
Following the widely-used setting in prior research \cite{DBLP:conf/cvpr/DongLPS0HL18, DBLP:conf/iclr/LinS00H20, DBLP:conf/iccv/WangH0021, DBLP:conf/cvpr/DongPSZ19}, we restrict perturbations in a $\ell_{\infty}$-norm ball in this paper. 
Therefore, the generation of adversarial examples can be formulated as an optimization problem:
\begin{equation}
\underset{\boldsymbol{x}^{adv}}{\arg \max}  \ 
    \mathcal{L}(\mathcal{M}_{\phi}({x}^{adv}), y), \quad \text { s.t. }\|{\delta}\|_{\infty} \leq \epsilon, \label{eq:attack}
\end{equation}
where $\mathcal{L}$ is the cross-entropy loss, which is commonly applied to the classification model. Nonetheless, in the black-box scenario, it is impossible to directly optimize Eq.~\eqref{eq:attack} because parameters of $\mathcal{M}_{\phi}$ are unknown. To address this issue, a common practice is to generate adversarial examples via a surrogate model $\mathcal{S}_{\theta}$ and mislead the target model by the transferability of adversarial examples. According to the I-FGSM~\cite{DBLP:conf/iclr/KurakinGB17}, adversarial example at $(t+1)$-th optimization iteration can be formulated as:
\begin{equation}
\label{eq:black_objective}
\boldsymbol{x}_{t+1}^{adv}=\operatorname{Clip}_{x}^{\epsilon}\{\boldsymbol{x}_{t}^{adv}+\alpha \cdot \operatorname{sign}(\nabla_{{x}_{t}^{adv}} \mathcal{L}(\mathcal{S}_{\theta}({x}_{t}^{adv}, y))))\},
\end{equation}
where $\operatorname{Clip}_{x}^{\epsilon}\{\cdot\}$ is an operation to constrain attack strength to be within a $\epsilon$-ball, $\alpha$ is the step size, $\operatorname{sign}(\cdot)$ denotes the sign function and $\nabla_{{x}_{t}^{adv}} \mathcal{L}(\mathcal{S}_{\theta}({x}_{t}^{adv}, y))$ represents the gradient of the loss. 
Moreover, if data transformation $\mathcal{T}(\cdot)$ is leveraged to boost the transferability of the adversarial examples, we can modify Eq.~\eqref{eq:black_objective} as a more unified paradigm:

\begin{equation}
\boldsymbol{x}_{t+1}^{adv}=\operatorname{Clip}_{x}^{\epsilon}\{\boldsymbol{x}_{t}^{adv}+\alpha \cdot \operatorname{sign}(\nabla_{{x}_{t}^{adv}} \mathcal{L}(\mathcal{S}_{\theta}(\mathcal{T}({x}_{t}^{adv}), y)))\}.
\label{eq:3}
\end{equation}

\subsection{Vanilla Warping Transformation (VWT)} \label{sec:vwt}

We have revealed in the Introduction that the warping transformation can achieve a greater diversity of local details and contents of data. 
Hence, we propose Vanilla Warping Transformation~(VWT) ($\mathcal{T}_{v}(x;\xi)$) controlled by random noise {map} $\xi$ to replace the $\mathcal{T}(\cdot)$ in Eq.~\eqref{eq:3}. 
To perform VWT, according to the core of TPS algorithm~\cite{bookstein1989principal, donato2002approximate}, we set two interpolation functions $\Phi_{x}$ and $\Phi_{y}$ for the coordinate offsets in the $x$-direction and $y$-direction, respectively. 
Then, two sets of control points are manually set to acquire the unknown parameter coefficients for the $\Phi_{x}$ and $\Phi_{y}$ functions. Specifically, the original control points are defined as $O \in \mathbb{R}^{M \times 2}$, where $M$ is the number of control points. The target control points ${P \in \mathbb{R}^{M \times 2}}$ are generated by slightly perturbing the original control points:
\begin{equation}
    P = O + \xi ,
\end{equation}
where $\xi \in \mathbb{R}^{M \times 2}$ are randomly sampled from a uniform distribution. Then, through these two group control points, the TPS coefficients of $\Phi_{x}$ and $\Phi_{y}$ are obtained. For brevity, we put more details of the TPS algorithm in the supplementary material. Finally, we are able to achieve the transformation $\mathcal{T}_{v}(x;\xi)$ by interpolation as:
\begin{equation}\label{EqVanillaDeform}
    \mathcal{T}_{v}(x_{t}^{adv}[m, n]; \xi) = x_{t}^{adv}[m+\Phi_{x}(m), n+\Phi_{y}(n)],
\end{equation}
 where $(m,n)$ is the coordinate of any pixel on $x_{t}^{adv}$, and $x_{t}^{adv}[m,n]$ denotes the pixel value corresponding to coordinate $(m,n)$. 
 We incorporate VWT into Eq. \eqref{eq:attack} and get an optimization objective to craft adversarial examples:
 \begin{equation}
    \underset{\|{\delta}\|_{\infty} \leq \epsilon} {\max } \ \mathcal{L} (\mathcal{S}_{\theta}(\mathcal{T}_{v}(x^{adv}_{t}; \xi), y)).
    \label{eq:vwt}
\end{equation}

\begin{figure}[tb]
    \centering
    \includegraphics[width=0.47\textwidth]{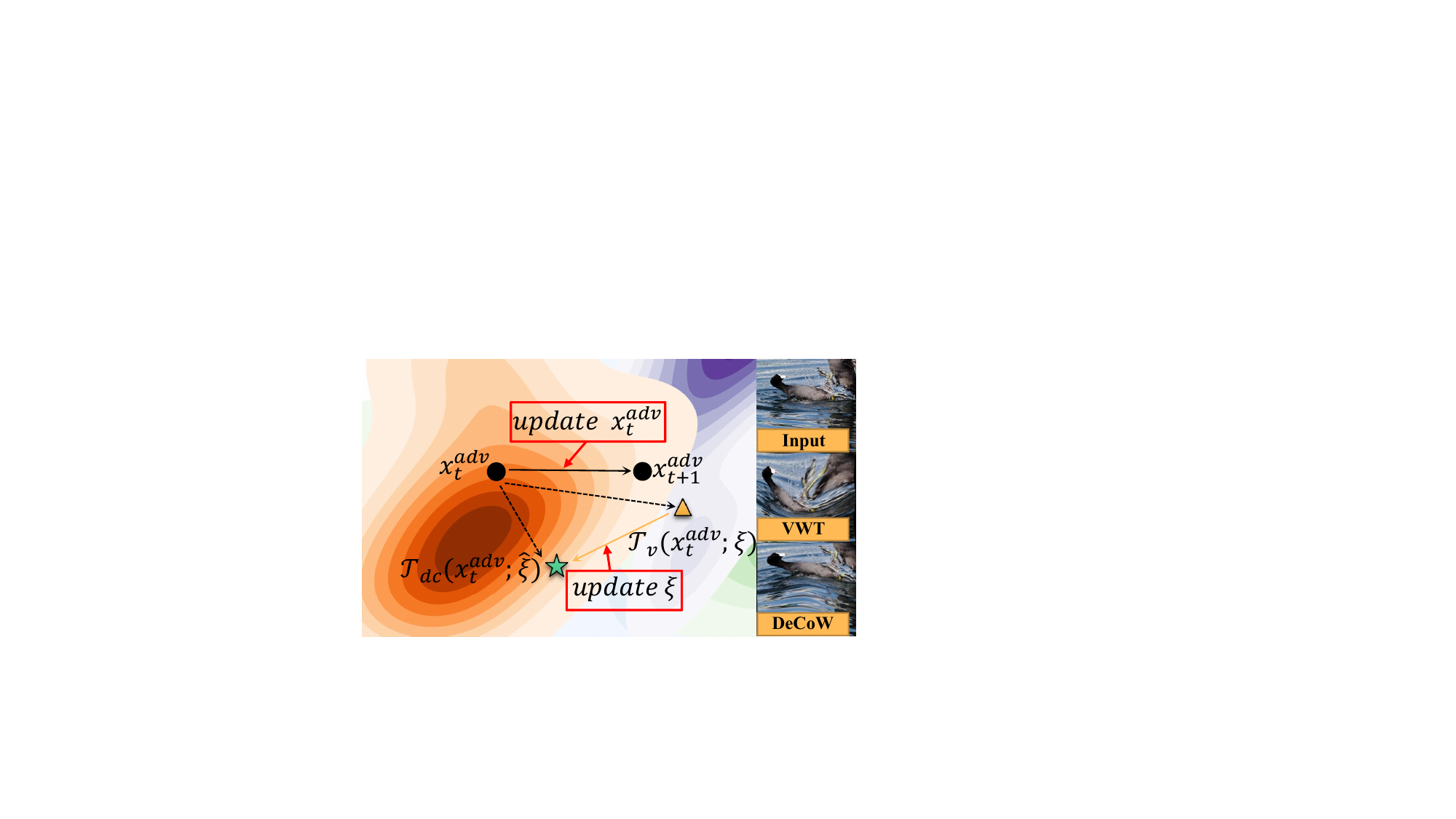}
    \caption{The process of updating $\xi$ and $x_{t}^{adv}$. The left part shows a diagram of the update process. The right column enumerates the input sample and its result after VWT and DeCoW, respectively.}

    \label{fig:method}
\end{figure}

\subsection{Deformation-Constrained Warping (DeCoW)} \label{sec:awt}

And yet, the magnitude and direction of VWT are dependent on a random noise map $\xi$ from a uniform distribution. 
Excessive and unreasonable deformation caused by $\xi$ may be posed on some image regions or video/audio segments, invalidating their global content and semantics.   
To constrain the deformation variable $\xi$,
we modify VWT with an adaptive control strategy.
We denote the new transformation, Deformation-Constrained Warping (DeCoW), as $\mathcal{T}_{dc}(x; \hat{\xi})$.
As illustrated in Figure~\ref{fig:method}, the content of the augmented sample with VWT is drastically changed while DeCoW can refine the deformation to an optimized point.

To achieve DeCoW, we first optimize initial point $\xi$ by minimizing an objective, which is a reverse optimization to the Eq.\eqref{eq:vwt}:
\begin{equation} \label{eq: arg min}
    \hat{\xi} =  \underset{\xi} {\arg \min } \ \mathcal{L} (\mathcal{S}_{\theta}(\mathcal{T}_{v}(x^{adv}_{t}; \xi)), y).
\end{equation}
Through this process, we can update random deformation noise $\xi$  to a secure point $\hat{\xi}$, reducing the variations of global semantics during elastic transformation.
More specifically, we implement an iterative update with back-propagated gradients to achieve the optimization, which is formulated as follows:
\begin{equation} \label{eq:update noise}
    \hat{\xi} = \xi - \beta \cdot \nabla_{\boldsymbol \xi}(\mathcal{L} (\mathcal{S}_{\theta}(\mathcal{T}_{v}(x^{adv}_{t}; \xi), y)),
\end{equation}
where $\beta$ is the learning rate. 
Based on this, we can get a new set of control points:
\begin{equation}
    P^{'} = O +\hat{\xi}.
\end{equation}

Consequently, two new interpolation functions $\hat{\Phi}_{x}'$ and $\hat{\Phi}_{y}'$ can be obtained to augment samples as follow:
    \begin{equation} \label{eq:get warping transform}
   \mathcal{T}_{dc}(x^{adv_{t}}[m, n]; {\hat{\xi}}) = x^{adv}_{t}[m+\hat{\Phi}_{x}'(m), n+\hat{\Phi}_{y}'(n)].
\end{equation}
Finally, we formulate this new attack loss as:
\begin{equation}
    \underset{\|{\delta}\|_{\infty} \leq \epsilon} {\max } \ \underset{\xi} {\min } \ \mathcal{L} (\mathcal{S}_{\theta}(\mathcal{T}_{v}(x^{adv}_{t}; \xi), y)).
\end{equation}

The defined max-min optimization can craft adversarial examples (maximization problem) with desire augmentation that has limited elastic deformation (minimization problem) to increase the diversity of local details and can keep image global semantics.

{\bf Warping for Video and Audio.} As a general approach, DeCoW can also be applied to augment data from other modalities (\emph{e.g.} video or audio). In processing these data with temporal information, we have made slight adjustments to allow it to be applied to the deformation of time series data. Specifically, suppose we want to apply DeCoW to a video $x \in \mathbb{R}^{K \times H \times W \times C}$ with $K$ frames. Here we first sample continuous random noise within a periodic function \emph{cosine function}, so that the noises have periodic prior information and relationships with each other. It can be formulated as follows:
\begin{equation}
    \xi_{\tau} = \{\xi^{(1)}_{\tau},\xi^{(2)}_{\tau},...,\xi^{(K)}_{\tau} \}.
\end{equation}
Here $\xi_{\tau} \in \mathbb{R}^{K \times M \times 2}$, $K$ is the number of video frames. And then, with this prior noise, we first perform an initial update on $\xi_{\tau}$ using Eq. \eqref{eq:update noise} and get $ \hat{\xi}_{\tau} = \{\hat{\xi}^{(1)}_{\tau},\hat{\xi}^{(2)}_{\tau},...,\hat{\xi}^{(K)}_{\tau} \}$. Subsequently, in order to achieve a more {smooth} noise between frames, we adopt a momentum accumulation fashion and perform another update on the $\hat{\xi}$ to strengthen the temporal correlation as:\
\begin{equation}\label{eq:temporal}
    \hat{\xi}^{(i+1)}_{\tau} = d \cdot \hat{\xi}^{(i)}_{\tau} + (1-d) \cdot \hat{\xi}^{(i+1)}_{\tau},
\end{equation}
where $d$ is a hyperparameter. Hence, through Eq. \eqref{eq:temporal}, we can perform continuous DeCoW for video clips and exploit the temporal information between consecutive frames.

\begin{table*} [!htbp]

 \adjustbox{max width=\linewidth}{
	\begin{tabular}{l|c|ccccccccc}
\hline
		\makecell[c]{\multirow{2}*{Surrogate}}  & Method  & ViT-B/16  &DeiT-B &LeViT-256 &PiT-B & CaiT-S-24 & ConViT-B &TNT-S  &Visformer-S & Avg\\
   \cline{2-11}
	  &Clean & 88.70 & 96.00 & 94.60 & 93.50  & 97.20 & 94.40 & 89.90 & 95.20 & 93.69  \\
\hline
   \multirow{7}*{ Inc-v3} & DIM &  64.80 &  69.30 &  53.90 &  66.90 &66.70 &70.80 &  47.40 & 55.10 & 61.86 \\
     & TIM & 66.20 &77.30 & 64.60 & 78.00 &76.50 & 78.50 & 54.00 & 70.00 & 70.64\\
    & SIM &  64.20 &  69.70 &  55.30 &  69.00 &67.10 & 70.40  &  47.80 & 59.60 & 62.89\\
   & Admix &  57.00 &  59.30 &  43.60 &  60.80 &57.60 & 63.10  &  37.60 & 46.40 & 53.34 \\
	 & S$^{2}$IM &  53.70 &  55.50 &  38.90 &  54.70  & 51.90 & 59.80 &  33.40 & 41.10 & \underline{48.62}  \\
          & DeCoWA & \textbf{44.80 } & \textbf{36.40} & \textbf{22.80}& \textbf{40.40} & \textbf{35.90} & \textbf{44.20} & \textbf{21.60} & \textbf{25.70} & \textbf{33.97} \\

\hline
     \multirow{7}*{ R50} & DIM & 60.30 & 53.00 & 39.80 & 47.50 &51.40 & 58.80  & 41.20 & 33.80 & 48.23 \\
     & TIM & 62.90 & 65.20 & 59.30 & 65.80 &66.00 & 68.00 & 50.40 & 55.20  & 61.60 \\
    & SIM  & 58.30 & 50.40 & 42.70 & 48.20 & 51.10 &55.00  & 42.40 & 34.50  & 47.82\\
   & Admix & 52.10 & 38.60 & 27.00 & 36.60 & 38.00 & 45.70 & 30.80 & 20.90 & 36.21\\
	 & S$^{2}$IM & 40.60 & 31.80 & 20.50 & 29.40 & 30.70 & 37.80  & 20.90 & {19.00} &\underline{28.84} \\
          &  DeCoWA & \textbf{37.90} & \textbf{23.50} & \textbf{13.90} & \textbf{20.60}& \textbf{23.50 }&\textbf{30.30 }& \textbf{15.20}& \textbf{11.00}& \textbf{21.99} \\

   \hline
	\end{tabular}}
	\caption{Classification accuracy (\%) against eight trained ViT models under the transferable adversarial attack with single input transformation, where all methods integrate MI-FGSM. 'Clean' indicates the accuracy before the attack. The best and the second-best performances are labeled in bold and {underline}, respectively. We abbreviate Inception-v3 to inc-v3 and abbreviate ResNet-50 to R50. 'Avg' is the average classification accuracy.} \label{tab:single-cnn}
\end{table*}

\subsection{Attack Algorithm} \label{sec:aa}
In this section, we proposed \textbf{Deformation-Constrained Warping Attack (DeCoWA)} algorithm by integrating DeCoW into MI-FGSM~\cite{DBLP:conf/cvpr/DongLPS0HL18} method. Firstly, every time we obtain the adversarial gradient $g^{'}$, a maximization-minimization operation is required, updating the random noise $\xi$ to obtain $\hat{\xi}$ via the minimization, and deriving the adversarial gradient $g^{'}$ based on $\hat{\xi}$ via the maximization. In addition, during the attack process, we apply multiple warping transformations to the adversarial example $x_{t}^{adv}$ as:
\begin{equation} \label{eq:final}
    \bar{g}_{t+1}=\frac{1}{N} \sum_{j=0}^{N} g_{j}^{'} =\frac{1}{N} \sum_{j=0}^{N} \nabla_{x_t^{adv}} \mathcal{L}( \mathcal{S}_{\theta}(
       \mathcal{T}_{dc}(x_t^{adv}; \hat{\xi}_{j})), y),
\end{equation} 
where $N$ is the number of adversarial warping transformations.
Multiple transformations allow us to enhance the model features in different directions, enhancing the diversity of surrogate models.
And then update the enhanced momentum:
\begin{equation} \label{eq:mom}
    g_{t+1} = \mu \cdot g_{t} + \frac{\bar{g}_{t+1}}{{\Vert \bar{g}_{t+1}  \Vert}_{1}},
\end{equation}
where $\mu$ denotes the decay factor.
Finally, based on the enhanced momentum $g_{t+1}$ , the adversarial example is updated as:
\begin{equation}  \label{eq:update}
{x}_{t+1}^{adv}=\operatorname{Clip}_{x}^{\epsilon}\{{x}_t^{adv}+\alpha \cdot \operatorname{sign}({g}_{t+1})\}.
\end{equation}

\section{Experiments}

\subsection{Attack on Image Classification} \label{exp:image}
In this section, we conduct extensive empirical evaluations on the trained image classifier and use the classification accuracy as the main evaluation metric, where the lower classification accuracy indicates better attack performance. Further details on the experimental setup and analysis are presented below.

\noindent {\bf Dataset.} Following the previous works~\cite{DBLP:conf/eccv/LongZZGLZS22}, we evaluate the proposed method on images from the ImageNet-compatible dataset\footnote{\url{https://github.com/cleverhans-lab/cleverhans/tree/master/cleverhans_ v3.1.0/examples/nips17_adversarial_competition/dataset}}.

\noindent {\bf Baseline.} We compare our approach with five state-of-the-art transfer-based attack methods, embracing Diverse Input Method
(DIM)~\cite{DBLP:conf/cvpr/XieZZBWRY19},  Translation-Invariant Method (TIM)~\cite{DBLP:conf/cvpr/DongPSZ19}, Scale-Invariant Method (SIM)~\cite{DBLP:conf/iclr/LinS00H20}, Admix~\cite{DBLP:conf/iccv/WangH0021} and Spectrum Simulation Attack Method (S$^2$IM)~\cite{DBLP:conf/eccv/LongZZGLZS22} . They also boost the transferability of adversarial examples by employing data augmentation. 
%
%
Note that MI-FGSM (MIM)~\cite{DBLP:conf/cvpr/DongLPS0HL18} is integrated into all the aforementioned methods.

\noindent {\bf Models.} To realize the cross model genus attack, two CNN models are chosen as the surrogate models, including Inception-v3~\cite{DBLP:conf/cvpr/SzegedyVISW16}, ResNet-50~\cite{DBLP:conf/cvpr/HeZRS16}. Then, the adversarial examples generated on CNN are tested on multiple ViT variants.
More comparative experiments are provided in the supplementary materials, including using two ViT models ViT-B/16~\cite{DBLP:conf/iclr/DosovitskiyB0WZ21} and DeiT-B~\cite{touvron2021training} to attack multiple CNN variants.

\noindent {\bf Attack Setting.} We follow the parameters setting~\cite{DBLP:conf/cvpr/DongLPS0HL18}. The perturbation budget is $\epsilon=16.0$, the number of iterations is $T=10$, and step size $\alpha = 1.6$. The decay factor for MIM is $\mu=1.0$. The Gaussian kernel size for TIM is $7\times7$. The number of copies is 5 for SIM. The transformation probability for DIM is $p=0.5$. The number of random samples for Admix is 3. In S$^2$IM, we set the number of spectrum transformations as 15. We set the number of DeCoW as $N=15$, the number of control points is $M=9$, learning rate $\beta=0.02$. 

\noindent {\bf Using CNNs to Attack ViTs.} In Table~\ref{tab:single-cnn}, we used CNN as the surrogate model to attack various variants of the ViT model, which is a challenging task mentioned in \cite{DBLP:conf/iccv/MahmoodMD21, shao2021adversarial, DBLP:conf/iccv/BhojanapalliCGL21}. Nevertheless, our proposed DeCoWA still achieves significant improvements. For instance, when applying Inception-v3 as the surrogate model, our attacking performance exceeds the existing best approaches S$^2$IM (48.62\%, underline) by over \textbf{14.65\%} on average.

\begin{table} [tb]
\centering
	
 \adjustbox{max width=\linewidth}{
	\begin{tabular}{l|c|cccc}
\hline
		\makecell[c]{\multirow{2}*{Surrogate}}  & Mehtod  & R101  & VGG19 & DN121  & EfficientNet\\
   \cline{2-6}
	  &Clean & 94.30  & 87.00 & 91.50  & 92.60    \\
   \hline
   \multirow{6}*{ Inc-v3} & DIM & 49.00  & 26.90& 31.40  & 31.80 \\
     & TIM  & 66.00  & 37.00 & 49.80  & 50.60 \\
    & SIM  & 53.80  & 30.90 & 33.30 &36.10 \\
   & Admix&39.00 & 21.90& 20.10  & 23.10  \\
	 & S$^{2}$IM &  36.20  &  18.80 &  19.30  &  21.40  \\
          & DeCoWA & \textbf{28.80}  & \textbf{12.80} & \textbf{11.60}  & \textbf{9.50} \\
   \cline{2-6}

   \hline
     \multirow{6}*{ R50} & DIM & 14.40 &22.80 & 17.60   & 27.20  \\
     & TIM& 32.30  & 32.30 & 34.70  & 41.40 \\
    & SIM  &12.80  & 26.10 &15.70 & 29.80 \\
   & Admix  & 5.80  & 16.40 & 8.40  & 17.10 \\
	 & S$^{2}$IM &  \textbf{5.40} &  9.20 &  7.10 &  12.80  \\
          &  DeCoWA & \textbf{5.40} &\textbf{6.80} & \textbf{3.90}& \textbf{7.80} \\

   \hline
	\end{tabular}}

 \caption{Classification accuracy (\%) against four trained CNN models under the transferable adversarial attack with single input transformation, where all methods integrate MI-FGSM. `Clean' indicates the accuracy before the attack. The best  performances are labeled in bold .} \label{tab:single-cnn-cnn}
\end{table}

\noindent {\bf Using CNNs to Attack CNNs.} In Table \ref{tab:single-cnn-cnn}, we realize a homologous model genera attack. It compares different adversarial examples generated from a CNN surrogate to other CNN targets (CNN $\rightarrow$ CNN). Here we have selected four classic CNN networks as the target models, i.e. ResNet-101~(R101), VGG19~\cite{simonyan2014very}, DenseNet121(DN121~\cite{huang2017densely}) and EfficientNet~\cite{tan2019efficientnet}. Under all settings, our DeCoWA achieved the best transferability as compared to existing state-of-the-art methods, which outperform the second-best method.  Generally speaking, the methods that achieve good results in cross model genus attacks can also perform well in homologous model genera attacks.

\begin{table} [!htbp]
\centering

 \adjustbox{max width=\linewidth}{
	\begin{tabular}{c|c|cccc}
\hline
	\makecell[c]{\multirow{1}*{Surrogate}}  & Method  & I3D  &SlowFast & TimeS &Swin-S \\
  \cline{1-6}
 \multirow{4}*{I3D-50} & BIM & 0.64$^{*}$ & 94.88 & 96.38 & 97.44 \\
  & DIM & 0.85$^{*}$ & 84.65 & 88.91 & 94.46\\
 & SIM & 0.00$^{*}$ & 73.56 &82.73 & 86.14 \\
 & DeCoWA & 0.43$^{*}$ & \textbf{70.58} & \textbf{79.32} & \textbf{83.37}\\
 \cline{1-6}
 \multirow{4}*{SlowFast} & BIM & 89.98 & 1.71 $^{*}$ & 94.88 & 98.72\\
  & DIM &80.81 & 2.35$^{*}$ & 85.29& 93.60 \\
 &SIM& {70.58} & 0.00$^{*}$ & 75.69 & 91.90\\
  & DeCoWA &\textbf{69.08 }& 2.13$^{*}$ &\textbf{73.35} & \textbf{88.49}\\
 \cline{1-6}

\multirow{4}*{TimeS} & BIM &88.49 & 91.26 & 0.00$^{*}$ & 84.01  \\
  & DIM & 69.51 & 73.99&0.21$^{*}$& 65.03\\
 &SIM& 55.22 & 53.94& 0.00$^{*}$ &  45.42\\
  & DeCoWA &\textbf{44.56} & \textbf{49.04} &0.00$^{*}$ & \textbf{39.23}\\

 \cline{1-6}
  \multirow{4}*{Swin-S} &BIM&92.54 & 94.03& 95.10&  0.00$^{*}$  \\
  & DIM&78.89 & 79.96&85.93 & 0.21$^{*}$\\
 &SIM& 63.11 &70.36&78.25&0.00$^{*}$\\
  & DeCoWA &\textbf{44.99}& \textbf{50.75} & \textbf{66.74}&0.00$^{*}$\\
\hline
	\end{tabular}}
 \centering
	\caption{Classification accuracy (\%) on four video recognition models. The adversaries are crafted on I3D-50, SlowFast, TimeSformer (TimeS), and Swin transformer. $^{*}$ indicates the attack performance under the white-box attack.} 

 \label{Tab:video}
\end{table}

\subsection{Attack on Video Recognition}
In this section, we show that our DeCoWA also can be easily applied in attacking the video recognition models.
\begin{table} [!tb]
\centering
	
\adjustbox{max width=\linewidth}{
	\begin{tabular}{c|c|cccc}
\hline

\makecell[c]{\multirow{2}*{Surrogate}}  & Method  & Baseline  &PANN & RGASC &ERGL \\
  \cline{2-6}
 & Clean & 68.30 & 73.40 & 77.40 & 75.50\\
 \cline{1-6}
\multirow{3}*{Baseline} & BIM& {0.10$^{*}$ }& 42.60 & 39.50 & 52.10\\
 & SI-FGSM & 13.90$^{*}$ & 46.80 & 43.80 & 54.80\\
 & DeCoWA &{ 0.20$^{*}$} & \textbf{40.80} & \textbf{37.60} & \textbf{50.80} \\
 \cline{1-6}
 \multirow{3}*{PANN} &BIM & 45.20 & {0.00$^{*}$} & 8.90 & 22.70\\
 &SI-FGSM& 55.00 & 35.70$^{*}$ & 39.90 & 43.80 \\
  & DeCoWA & \textbf{44.10}& {0.00$^{*}$} & \textbf{ 7.90} &\textbf{ 21.40} \\
\cline{1-6}
\multirow{3}*{RGASC} & BIM &33.30 &57.50 &17.90$^{*}$ &48.10 \\
 &SI-FGSM&37.40 & 57.90&24.30$^{*}$ & 49.30 \\
  & DeCoWA &\textbf{28.70}&\textbf{55.50}&{13.90$^{*}$} & \textbf{44.50} \\

\cline{1-6}
   \multirow{3}*{ERGL} &BIM & 61.20 & 46.70 & 45.50 & 0.60$^{*}$ \\
 &SI-FGSM & 63.00 & 54.80 & 55.30 & 32.40$^{*}$\\
  & DeCoWA & \textbf{59.50} & \textbf{42.10} & \textbf{39.90} & {0.40$^{*}$} \\
\hline
	\end{tabular}}
 \caption{ Classification accuracy (\%) on four audio recognition models. The adversaries are crafted on Baseline, PANN, ERGL, and RGASC, respectively. $^{*}$ indicates the performance under the white-box attack. } 

\label{Tab:audio}

\end{table}

\noindent {\bf Attack Setting.} We evaluate our approach using Kinetics-400~\cite{DBLP:journals/corr/KayCSZHVVGBNSZ17} (K400) datasets, which are widely used for action video recognition. 469 videos are chosen from the validation set to evaluate the effectiveness of our algorithm. Our proposed method is evaluated on four action video recognition models, i.e. I3D~\cite{DBLP:conf/cvpr/0004GGH18}, SlowFast~\cite{DBLP:conf/iccv/Feichtenhofer0M19}, TimesFormer~\cite{DBLP:conf/icml/BertasiusWT21} and Swin Transformer~\cite{DBLP:conf/iccv/LiuL00W0LG21}. All the models are trained on Kinetics-400. The spatial size of the input is 224$\times$224. We make modifications based on the mmaction\footnote{\url{https://github.com/open-mmlab/mmaction2}} to implement the attack for the frames. 
We skip every three frames to select 32 consecutive frames to construct an input clip, and we get 3 input clips for each video. We evaluate the performances against three popular transfer-based attacks, i.e. BIM~\cite{DBLP:conf/iclr/KurakinGB17}, DIM~\cite{DBLP:conf/cvpr/XieZZBWRY19} and SIM~\cite{DBLP:conf/iclr/LinS00H20}.

\noindent {\bf Experiment Analysis.} Table \ref{Tab:video} shows the comparison results, it shows that our DeCoW can process consecutive frames in a temporal manner, outperforming other input transformation methods in attacking video recognition models. This demonstrates the generality of our method. Meanwhile, we observed that our method has a more pronounced augmentation effect on ViT models and achieves stronger attack performance when used as a substitute model.

\subsection{Attack on Audio Recognition}

In this section, we show that our DeCoWA also can be easily applied in attacking the audio recognition models.

\noindent {\bf Attack Setting.} Four acoustic scene classification models, i.e. Baseline\footnote{\url{https://github.com/qiuqiangkong/dcase2018_task1}}, PANN~\cite{kong2020panns}, ERGL~\cite{DBLP:journals/corr/abs-2210-15366} and RGASC~\cite{DBLP:conf/ijcnn/HouKHB22}, and 2,518 audios selected from the validation set are used for the evaluation.
We compared the proposed DeCoWA against two popular transfer-based attacks, i.e. BIM~\cite{DBLP:conf/iclr/KurakinGB17} and SI-FGSM~\cite{DBLP:conf/iclr/LinS00H20}.
 All the models are trained on TUT Urban Acoustic Scenes 2018. For this comparison, the image transformation method of DIM is not used, since it can not {be applied} to process speech signals directly, while our method can process speech signals easily. 

\begin{figure}[tb]
    \centering
    \includegraphics[width=0.47\textwidth]{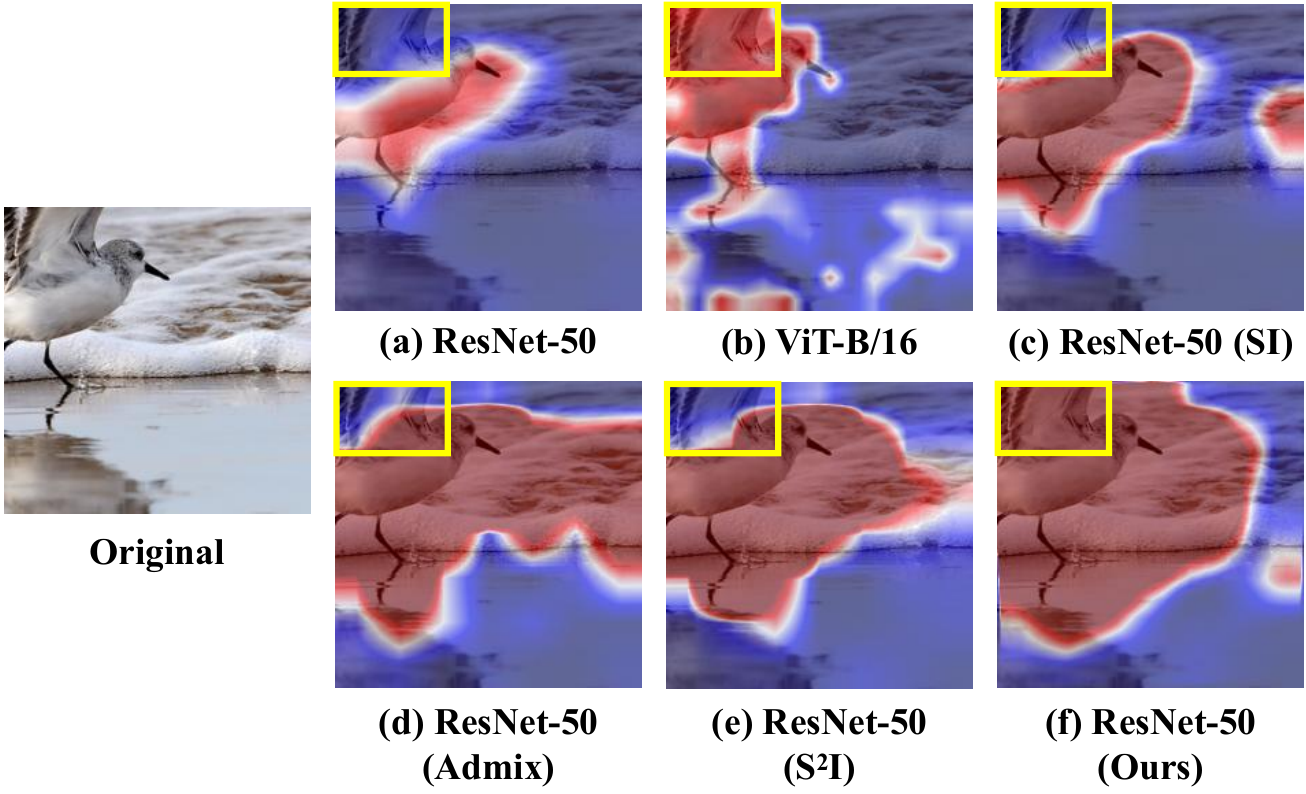}

    \caption{Visualization of Grad-CAM \cite{DBLP:conf/iccv/SelvarajuCDVPB17} for two trained models ResNet-50 and ViT-B/16. (a)$\sim$(b): the results for raw images on ResNet-50 and ViT-B/16. (c)$\sim$(e): the results for SI ~\cite{DBLP:conf/iclr/LinS00H20}, Admix ~\cite{DBLP:conf/iccv/WangH0021}, S$^{2}$I~\cite{DBLP:conf/eccv/LongZZGLZS22} images on ResNet-50. (f): the result for our DeCoW images on ResNet-50.}

    \label{fig:gcam}
\end{figure}

\noindent {\bf Experiment Analysis.} Table \ref{Tab:audio} shows that our DeCoWA consistently outperforms the state of the arts by crafting more generalized perturbations. Meanwhile, as far as we know, there are currently few methods aiming to specifically improve the transferability of audio adversarial samples, our algorithm provides a new approach to attack such systems. Note that here our DeCoW only combines with I-FGSM.

\subsection{Visualization of Grad-CAM}

To shed light on how our method works, we visualize the Grad-CAMs by ResNet-50, ViT-B in Figure~\ref{fig:gcam}. 
As shown in Figure~\ref{fig:gcam} (a)$\sim$(b) and (f), ResNet-50, which is prone to focus more on local and sparse regions of an object, can be transformed by our DeCoW to recognize an object in terms of its global appearance. For example, in the yellow box, DeCoW makes ResNet-50 pay attention to the bird's wing as well like the way of ViT-B/16, which enables the surrogate ResNet-50 to simulate the ViT-B/16 successfully. Still, other augmentation methods fail to achieve it. This indicates that the proposed DeCoW is a generalized transform and can explore more attention areas of the target system, and consequently better narrow the gap between CNNs and ViTs. 
Noticed that here we apply the CAM to the original sample instead of the enhanced sample to show how the augmented sample shift and extend the original CAM (Figure~\ref{fig:gcam}(a)). As a result, the backgrounds in Figure~\ref{fig:gcam} are similar.

\begin{figure}[tb]
     \centering
    \footnotesize
    \includegraphics[width=1.0\linewidth]{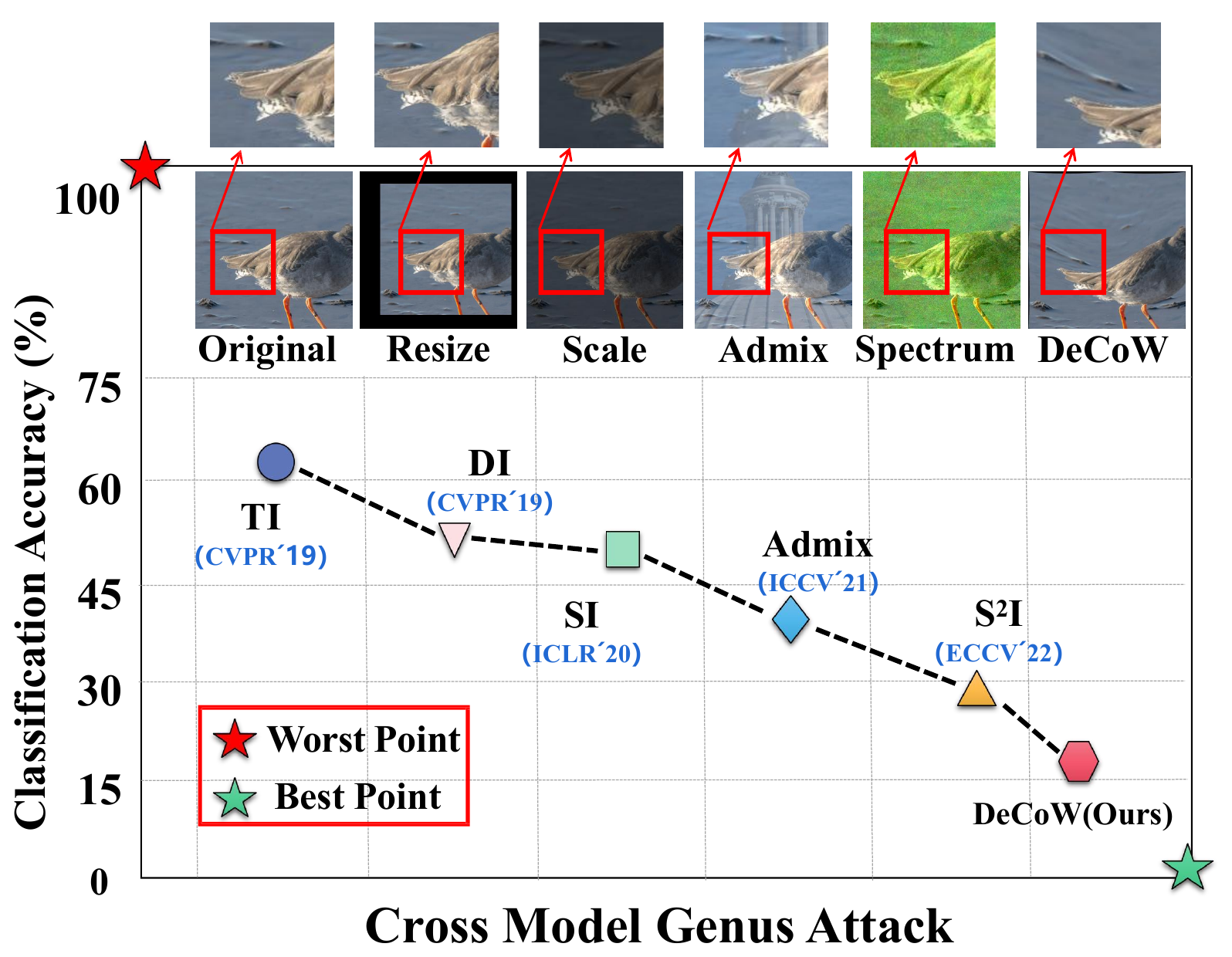}
     \caption{In comparison with other input transformation methods, our method makes profound changes with the local shape and contours (red box) thus accessing diverse localities, while others can only increase global diversity.}
     \label{fig:motivation2}
 \end{figure}

\subsection{Visualization of Different Input Transformation}

We present prior augmentation methods and their performance in Figure~\ref{fig:motivation2}.
Such methods augment input samples from various perspectives, including 
image size~\cite{DBLP:conf/cvpr/XieZZBWRY19}, translation~\cite{DBLP:conf/cvpr/DongPSZ19}, 
scale~\cite{DBLP:conf/iclr/LinS00H20}, linear multiple-image fusion~\cite{DBLP:conf/iccv/WangH0021} and spectrum~\cite{DBLP:conf/eccv/LongZZGLZS22}.
These transformations are prone to change global contents (\emph{e.g.,} size, position, lighting, or color), while we seek a new transformation to preserve global semantics and increase the diversity of local details like local shape and contours, which are more general and invariant features~\cite{DBLP:conf/iccv/MahmoodMD21, DBLP:conf/iclr/GeirhosRMBWB19} to both CNN and Transformer-based models.
Figure~\ref{fig:motivation2} depicts that warping transformation can cause deformation to the local region (red box), while the others are only able to make limited changes on the tail of the bird (lack of diversity).
\section{Conclusion and Outlook}
In this work, we argue that more attention should be paid to the task of cross model genus attacks. We proposed a novel technique, Deformation-Constrained Warping Attack (DeCoWA) to boost the adversarial transferability across model genera. It features applying constrained elastic deformation to input samples to simulate diverse models covering different model genera. 
Comprehensive experiments corroborate the superiority of DeCoWA for the task of cross model genus attack on data of various modalities. Therefore, our attack can serve as a strong
baseline to compare future cross model genus attack. 
In the future, we will focus on the cross-data distribution attack in which the adversary can only access surrogate models trained on different data distributions and having distinct model genera from the target system.

\label{sec:reference_examples}

\section{Acknowledgments}
 The work was supported by the Natural Science Foundation of China under grants no. 62276170, 82261138629, 
the Science and Technology Project of Guangdong Province under grants no. 2023A1515011549, 2023A1515010688,
 the Science and Technology Innovation Commission of Shenzhen under grant no. JCYJ20220531101412030.

\bigskip
\bibliography{aaai24}

\end{document}